\documentclass[acmtog,nonacm,balance=false]{acmart}
\AtBeginDocument{%
  \providecommand\BibTeX{{%
    \normalfont B\kern-0.5em{\scshape i\kern-0.25em b}\kern-0.8em\TeX}}}

\setcopyright{none}

\newcommand{\revRemoved}[1]{{}}

\newcommand{\eg}{\emph{e.g.}} 
\newcommand{\cf}{\emph{cf.\ }} 
\newcommand{\vs}{\emph{vs.}} 

\renewcommand{\eqref}[1]{Equation~\ref{eq:#1}}

\newcommand{\tabref}[1]{Table~\ref{tab:#1}}

\newcommand{\MethodNameFull}{\textsc{RefFusion}~}
\newcommand{\MethodName}{\textsc{RefFusion}~}

\usepackage{multirow}
\usepackage{enumitem}
\usepackage{amsmath}
\usepackage{amsfonts}
\usepackage{nicefrac}
\usepackage{mathtools}
\usepackage{wrapfig}
\usepackage{tikz}
\usepackage{xcolor}
\usepackage{colortbl}
\usepackage{adjustbox}
\usepackage{booktabs} 
\usepackage{tabularx}
\usepackage{xspace}
\usepackage{ulem}
\newcolumntype{Y}{>{\centering\arraybackslash}X}
\newcolumntype{C}[1]{>{\centering\arraybackslash}p{#1}}

\usepackage{xcolor}

\usepackage{custom_style} 

\usepackage{arydshln}

\usepackage{amsmath,amsfonts,bm}

\def\Figref#1{Figure~\ref{#1}}

\def\Secref#1{Section~\ref{#1}}

\def\eqref#1{equation~\ref{#1}}
\def\Eqref#1{Equation~\ref{#1}}

\def\1{\bm{1}}

\def\rvepsilon{{\bm{\epsilon}}}

\def\rvc{{\mathbf{c}}}

\def\rvx{{\mathbf{x}}}

\def\vzero{{\bm{0}}}

\def\vtheta{{\bm{\theta}}}
\def\vpi{{\bm{\pi}}}

\def\vc{{\bm{c}}}

\def\vm{{\bm{m}}}

\def\vr{{\bm{r}}}
\def\vs{{\bm{s}}}

\def\vx{{\bm{x}}}
\def\vy{{\bm{y}}}
\def\vz{{\bm{z}}}

\def\mI{{\bm{I}}}

\DeclareMathAlphabet{\mathsfit}{\encodingdefault}{\sfdefault}{m}{sl}
\SetMathAlphabet{\mathsfit}{bold}{\encodingdefault}{\sfdefault}{bx}{n}

\def\gD{{\mathcal{D}}}
\def\gE{{\mathcal{E}}}

\def\gG{{\mathcal{G}}}

\def\gN{{\mathcal{N}}}

\def\sZ{{\mathbb{Z}}}

\newcommand{\px}{p(\mathbf{x})}

\newcommand{\E}{\mathbb{E}}
\newcommand{\Ls}{\mathcal{L}}

\newcommand{\dmparam}{\vtheta}
\newcommand{\denoiser}{\rvepsilon_{\dmparam}}
\newcommand{\gparam}{\bm{\phi}}
\newcommand{\gnet}{g_{\gparam}}
\newcommand{\encoder}{\gE}

\newcommand{\encodertext}{\gE_{\textnormal{text}}}

\newcommand{\imgRendered}{{\hat {\mathbf{I}}}}
\newcommand{\imgReference}{{\mathbf{I}_r}}
\newcommand{\imgReferenceCropped}{{\mathbf{I}_r^c}}

\newcommand{\imgPatchRendered}{\hat{\mathbf{I}}^p}

\newcommand{\globalText}{{\textit{"A photo of sks"}}}
\newcommand{\localText}{{\textit{"A photo of sks, cropped"}}}

\begin{document}

\title{RefFusion: Reference Adapted Diffusion Models for 3D Scene Inpainting}

\author{Ashkan Mirzaei}
\affiliation{%
  \institution{NVIDIA, University of Toronto}
  \country{Canada}
}
\email{ashkan@cs.toronto.edu}

\author{Riccardo de Lutio}
\affiliation{%
  \institution{NVIDIA}
  \country{USA}
}
\email{rdelutio@nvidia.com}

\author{Seung Wook Kim}
\affiliation{%
  \institution{NVIDIA}
  \country{South Korea}
}
\email{seungwookk@nvidia.com}

\author{David Acuna }
\affiliation{%
  \institution{NVIDIA}
  \country{Canada}
}
\email{dacunamarrer@nvidia.com}

\author{Jonathan Kelly}
\affiliation{%
  \institution{University of Toronto}
  \country{Canada}
}
\email{jkelly@utias.utoronto.ca}

\author{Sanja Fidler}
\affiliation{%
  \institution{NVIDIA, University of Toronto}
  \country{Canada}
}
\email{sfidler@nvidia.com}

\author{Igor Gilitschenski}
\affiliation{%
  \institution{University of Toronto}
  \country{Canada}
}
\email{igor@gilitschenski.org}

\author{Zan Gojcic}
\affiliation{%
  \institution{NVIDIA}
  \country{Switzerland}
}
\email{zan.gojcic@gmail.com}

\acmSubmissionID{1017}


\begin{abstract}

Neural reconstruction approaches are rapidly emerging as the preferred representation for 3D scenes, but their limited editability is still posing a challenge. In this work, we propose an approach for 3D scene inpainting---the task of coherently replacing parts of the reconstructed scene with desired content. Scene inpainting is an inherently ill-posed task as there exist many solutions that plausibly replace the missing content. A good inpainting method should therefore not only enable high-quality synthesis but also a high degree of control.
Based on this observation, we focus on enabling explicit control over the inpainted content and leverage a reference image as an efficient means to achieve this goal. Specifically, we introduce \MethodNameFull, a novel 3D inpainting method based on a multi-scale personalization of an image inpainting diffusion model to the given reference view. The personalization effectively adapts the prior distribution to the target scene, resulting in a lower variance of score distillation objective and hence significantly sharper details. Our framework achieves state-of-the-art results for object removal while maintaining high controllability. We further demonstrate the generality of our formulation on other downstream tasks such as object insertion, scene outpainting, and sparse view reconstruction.
 
\end{abstract}

\begin{teaserfigure}
\centering
\includegraphics[width=1\textwidth]{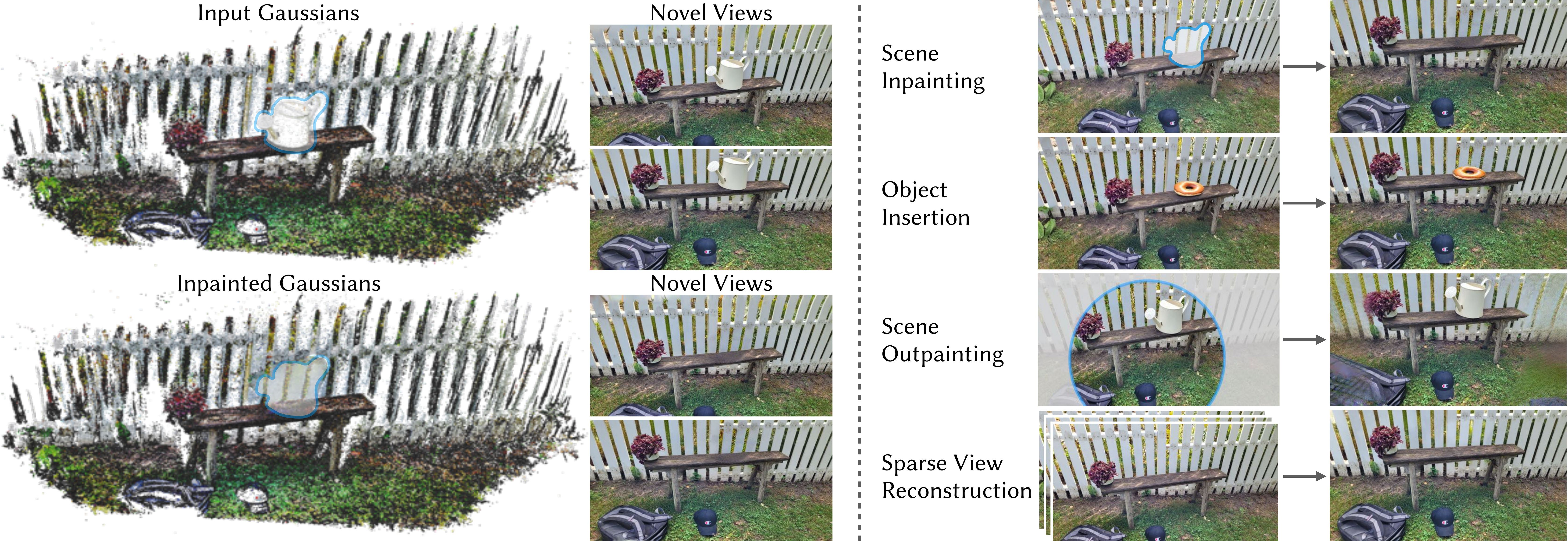}
\caption{
This work presents an approach for 3D inpainting based on the distillation of the 2D generative priors from reference adapted diffusion models. \emph{Left} Given a scene represented using 3D Gaussian splatting and the user-defined mask, we use the score distillation objective based on a personalized diffusion model to inpaint the missing content in 3D. \emph{Right} Our method is general and can be, without any changes, applied to other editing tasks such as object insertion, outpainting, and sparse view reconstruction. Please visit our \href{https://reffusion.github.io}{\color{magenta}{project page}}.} 
\end{teaserfigure}

\maketitle

\section{Introduction}

Neural reconstruction methods enable seamless reconstruction of 3D scenes from a set of posed images. Their simple formulation, high visual fidelity, and increasingly fast rendering make them the preferred representation for a variety of use cases from AR/VR applications to robotics simulation. While these methods enable novel-view synthesis, they are still inherently limited to the content captured in the training data. Yet, to make them useful in practice, it is critical to impart them with editability. One of the key desirable manipulation operations is the ability to remove parts of the scene and substitute them with some desired content. This task, generally known as \emph{3D inpainting}, involves synthesizing plausible content that coherently blends with the rest of the scene when viewed from any angle. Inpainting is an inherently ill-posed problem as there often exist multiple viable approaches to complete the scene. A good inpainting model should therefore also be controllable, such that users can choose the solution that best fits their needs.

\begin{figure*}
    \centering
    \includegraphics[width=\textwidth]{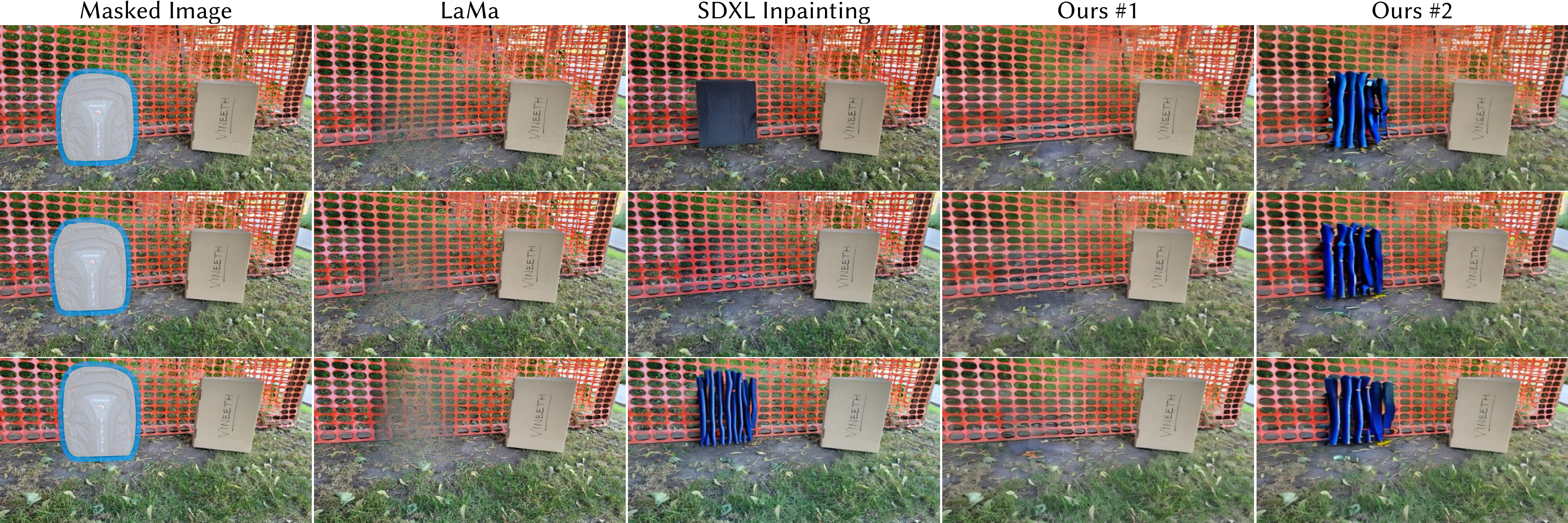}
    \vspace{-6mm}
    \caption{Comparison of 2D image inpainting methods on multiple views of the same scene. LaMa~\cite{suvorov2022resolution} yields relatively consistent inpaintings but lacks details. SDXL~\cite{podell2023sdxl} synthesizes content with high-quality, but low multi-view consistency due to the high diversity of its generations. By personalizing the diffusion model to the reference view, our method achieves high-quality generations with superior multi-view consistency. \emph{Ours \#1} and \emph{Ours \#2} are adapted to SDXL outputs shown in the second and third row respectively.}
    \label{fig:motivation_tradeoff}
    \vspace{-2mm}
\end{figure*}
With the advent of large diffusion models \cite{rombach2021highresolution, balaji2022ediffi, saharia2022imagine, saharia2022palette}, 2D image inpainting has been getting increasingly close to achieving these properties. Unfortunately, large-scale 3D generative methods still lag far behind. This can be accredited to the lack of large-scale 3D datasets and efficient high-resolution 3D data structures. As a consequence, much of the recent and concurrent work on 3D inpainting has resorted to lifting the priors of 2D inpainting models to 3D. Such lifting is performed either \emph{explicitly}, by independently inpainting one or multiple 2D views and consolidating their multi-view inconsistencies in 3D~\cite{liu2022nerfin, mirzaei2023spin, mirzaei2023reference, wang2023inpaintnerf360}. Alternatively, it can be formulated as a \emph{continuous distillation} process~\cite{poole2022dreamfusion} that seeks 3D inpaintings using a 2D diffusion model as a prior for optimization~\cite{prabhu2023inpaint3d}.

\begin{figure*}
    \centering
   \includegraphics[width=0.98\textwidth]{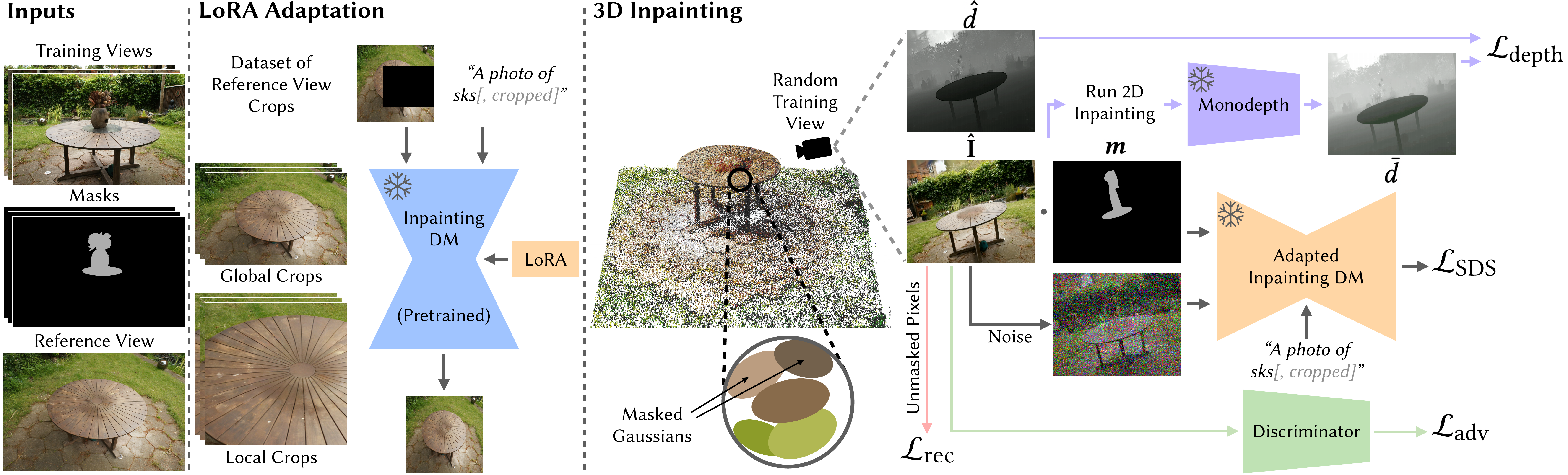}
    \caption{Overview of the proposed approach. \MethodName takes training views, masks, and the reference view as input (\emph{left}). We adapt the inpainting LDM on both the global and local crops of the reference view (\emph{middle}). Then, we distill the priors of the adapted LDM to the scene (\emph{right}) by minimizing the SDS objective. Additionally, we use a discriminator loss to mitigate potential artifacts in appearance and a depth loss to enhance geometry. We track Gaussians representing the masked and unmasked regions, and backpropagate the gradients of individual terms only to the pertinent regions.}
    \label{fig:method_diagram}
    \vspace{-2mm}
\end{figure*}

These methods achieve promising results but many challenges remain: \textbf{a)} \emph{Trade-off between diversity and multi-view consistency.} Deterministic 2D inpainting models, \eg~\cite{suvorov2022resolution}, yield relatively multi-view consistent inpaintings that can directly be lifted to 3D using perceptual losses~\cite{mirzaei2023spin, wang2023inpaintnerf360}. Yet, this comes at the cost of diversity and limited visual quality. Prompt-based inpainting diffusion models can synthesize high-quality, diverse, and controllable inpaintings, but due to the loose constraint of text guidance, their inpaintings are highly multi-view inconsistent and thus difficult to lift to 3D (see~\Figref{fig:motivation_tradeoff}). 
\textbf{b)} \emph{Maintaining fidelity to the observed content.} By inpainting the 3D content using (inconsistent) 2D masks, these methods ignore that the missing content in 3D is potentially smaller than that naively defined by the 2D masks. For example, in the scene shown in \Figref{fig:method_diagram}, the content behind the vase is observed in other views and therefore does not need to be synthesized.  
Finally, \textbf{c)} \emph{Conflicting gradients.} The inconsistencies across the inpainted views may lead to conflicting gradients, which in turn result in smoothed-out inpaintings.

To address the above challenges, we propose \MethodNameFull, a novel 3D inpainting method based on \emph{continuous distillation} of a reference adapted diffusion model. In particular: \textbf{i)} We propose a multi-scale personalization method for inpainting diffusion models that adapts the model to the given reference view. This contribution alone enables explicit control over the inpainted content and reduces the variance of the score distillation objective. \textbf{ii)} We leverage the explicit nature of 3D Gaussian splatting to consolidate noisy 2D masks and direct the gradients of different loss terms to the pertinent regions. \textbf{iii)} We propose a combination of objective terms that enables using the SDS optimization procedure at the scene level. \textbf{iv)} We propose a new dataset designed for evaluating object removal and 3D inpainting, comprising nine scenes with large camera motion.

The experiments in \Secref{sec:experiments} demonstrate the high visual quality, controllability, and diversity of our approach. In addition, the applications in \Secref{sec:applications} showcase the generality of our formulation.

\section{Related work}
\noindent\textbf{2D and 3D Inpainting} Inpainting is the process of replacing missing regions with realistic content. For example, in 2D this involves producing plausible values for missing pixels of an image. As an inherently generative task, advancements of generative models have led to increased performance of 2D inpainting. Generative Adversarial Network (GAN)~\cite{goodfellow2014generative}-based approaches~\cite{liu2021pd,zhao2021large,zheng2022image,yu2019free,li2022mat,suvorov2022resolution} learn to hallucinate the missing pixels by playing an adversarial game between an image inpainter and a discriminator.
Recently, diffusion model (DM)~\cite{ho2020ddpm,song2020score,sohl2015deep}-based inpainting models~\cite{rombach2021highresolution,xie2023smartbrush,avrahami2022blended,lugmayr2022repaint,meng2021sdedit} have achieved state-of-the-art results. These operate by gradually perturbing a clean image towards random noise while training a denoising network to reconstruct the image, conditioned on its masked version.

In 3D inpainting, the goal is to synthesize plausible content for the missing regions in 3D scenes. This is a significantly harder task as the generated content needs to be consistent across various views. Although 3D generative models have recently gathered increased interest \cite{nichol2022pointe,zeng2022lion,kim2023nfldm,bautista2022gaudi,kalischek2022tetrahedral}, their quality is still limited by the scarcity of 3D training data and the difficulty of selecting an appropriate underlying representation. 
Therefore, most existing 3D inpainting models~\cite{mirzaei2023spin,mirzaei2023reference,wang2023inpaintnerf360,weder2023removing,instructnerf2023,liu2022nerfin,prabhu2023inpaint3d,weber2023nerfiller} still rely on lifting the priors from 2D inpainting models to 3D. In particular, they first inpaint masked input images and complement the reconstruction objective with regularization techniques to lessen the multi-view inconsistencies~\cite{mirzaei2023spin,mirzaei2023reference,weder2023removing,wang2023inpaintnerf360,liu2022nerfin}. 
The problem of inconsistency is amplified when using recent DM-based inpainting models with increased diversity.
We address the challenge with the score distillation objective~\cite{poole2022dreamfusion} as follows.




\noindent\textbf{Distilling 2D Diffusion Model Priors} Score Distillation Sampling (SDS), first introduced by \cite{poole2022dreamfusion}, has recently been used to generate realistic 3D objects~\cite{lin2023magic3d,chen2023fantasia3d,wang2023prolificdreamer,liang2023luciddreamer} and even 4D scenes~\cite{ling2023alignyourgaussians,ren2023dreamgaussian4d,zheng2023unified} by distilling the priors of text-to-image DMs~\cite{rombach2021highresolution,dai2023emu,saharia2022imagine,balaji2022ediffi}. 
The SDS formulation for 3D scenes works by backpropagating gradients from a DM's denoiser to the underlying scene representation so that the renderings look realistic. 
By running this optimization process over many camera views this results in a 3D consistent scene representation.


In this paper, we propose to follow this paradigm and use a pretrained 2D inpainting DM as a prior to guide the optimization process using SDS.
Concurrent works Inpaint3D~\cite{prabhu2023inpaint3d} and NeRFiller~\cite{weber2023nerfiller} operate on a similar scheme, either by using the SDS objective directly or its iterative dataset update formulation proposed by IN2N~\cite{instructnerf2023}. 
We find that personalizing the DM to the target scene and reference image is key to achieving sharp results and user controllability. 


\noindent\textbf{Personalizing Diffusion Models}
Adapting large-scale text-to-image DMs to generate user-specified content offers an efficient way for obtaining a high-quality personalized generative model. This principle has been successfully applied to many applications, such as preserving the identity of an object~\cite{ruiz2022dreambooth} or rendering a chrome ball to estimate lighting~\cite{phongthawee2023diffusionlight}. 
Different personalization approaches have been proposed; DreamBooth~\cite{ruiz2022dreambooth} finetunes a DM and Textual inversion~\cite{gal2022image}  optimizes a new word embedding that can generate targets through a pretrained DM.
LoRA~\cite{hu2021lora,shah2023ziplora}, originally proposed for language models, injects trainable rank-decomposed matrices into frozen DMs for parameter-efficient personalization. 
RealFill~\cite{tang2023realfill} and \cite{chari2023personalized} personalize both the context encoder and DM for target images. 

Personalized DMs have been previously used for 2D to 3D distillation. ProlificDreamer~\cite{wang2023prolificdreamer} and DreamCraft3D~\cite{sun2023dreamcraft3d} use LoRA to learn an evolving model and compute a modified SDS objective for text-to-3D generation. Similarly, we personalize a pretrained DM using LoRA, but we propose a robust pipeline tailored for 3D inpainting to efficiently personalize the DM on an inpainted reference view and use it to inpaint a scene by optimizing an SDS objective.

\vspace{-5mm}
\section{Method}

Our method improves the distillation of 2D DM priors for 3D inpainting of scenes represented using Gaussian splatting~\cite{kerbl20233Dgaussians}. To this end, we adapt a 2D inpainting DM by employing multi-scale crops derived from a reference image~(\Secref{sec:multiscale_personalization}). The adaptation largely reduces the variance of the score distillation objective and removes the need for text guidance. Moreover, it allows us to introduce a multi-scale score distillation objective (\Secref{sec:multiscale_sds}) that encompasses both global context and local details. To guide the supervision of the reconstruction to the pertinent regions, we leverage the explicit nature of the Gaussian representation~(\Secref{sec:splitting_the_gaussians}). Important preliminaries are summarized in ~\Secref{sec:preliminaries}, while the implementation details are provided in~\Secref{sec:implementation_details}.

\subsection{Preliminaries}
\label{sec:preliminaries}

\noindent\textbf{Diffusion Models} are a family of generative models known for their ability to transform samples from a tractable distribution, typically a Gaussian, towards the target data distribution $\px$ \cite{sohl2015deep,ho2020ddpm}. These models are built around two key processes. A forward process, which gradually removes the structure from the data samples $\rvx \sim \px$ by adding noise. And, a reverse process, which slowly removes this noise and reintroduces the structure into an intermediate latent variable denoted as $\rvx_t=\alpha_t \rvx + \sigma_t \rvepsilon, \rvepsilon \sim \gN(\vzero,\mI)$. In this context,  $\alpha_t$ and $\sigma_t$ represent predetermined noise schedules, and the variable $t$ denotes the timestep with a higher value indicating a greater amount of noise. The reverse process is typically parameterized by a conditional neural network $\denoiser$ that is trained to predict the noise $\rvepsilon$  using the following simplified objective  \cite{ho2020ddpm}:
\begin{equation}
    \E_{\rvx \sim \px,\rvepsilon \sim \gN(\vzero,\mI), t \sim T}[w(t)|| \denoiser(\rvx_t,t,\rvc) -\rvepsilon ||^2_2],
\label{eq:vanilla_difussion}
\end{equation}
where $\vc$ represents a condition (\eg~text, image, etc.) that allows controlling the generation process, $w(t)$ represents a time-conditional weighting, and $T$ is a set containing a selection of timesteps.

Latent Diffusion Models (LDMs), improve computational and memory efficiency over traditional DMs by performing the diffusion process in a lower dimensional latent space~\cite{rombach2021highresolution}. This dimensionality reduction is achieved by employing a pretrained encoder–decoder architecture, where the encoder $\gE$ maps samples from the data distribution $\rvx \sim \px$ into a latent space $\sZ$. The decoder $\gD$ performs the inverse operation, such that $\gD(\gE(\vx)) \approx \vx$. In LDMs, the DM operates on $\sZ$, therefore $\vx$ in \Eqref{eq:vanilla_difussion} is replaced by its latent representation $\vz:=\gE(\vx)$.

\noindent\textbf{Score Distillation Sampling} originally proposed by \cite{poole2022dreamfusion}, leverages a pretrained DM to guide the optimization of a differentiable, parametric image-rendering function $\gnet(\vpi):=\vx$, where $\vpi$ denotes the camera pose from which the image $\vx$ is rendered. Specifically, the parameters $\gparam$ are updated using the 
gradient:
\begin{equation}
\label{eq:sds_gradient}
\nabla_{\gparam} \Ls_{\text{SDS}}(\gparam,\dmparam):= \E_{\rvepsilon \sim \gN(\vzero,\mI), t \sim T}[w(t) (\hat{\rvepsilon}_{\dmparam}(\vz_t,t,\vc)-\rvepsilon) \frac{\partial \vz_t}{\partial \gparam} ],
%
\end{equation}
where $\vz = \gE(\gnet(\vpi))$ and 

\begin{equation}
\hat{\rvepsilon}_{\dmparam}(\vz_t,t,\vc):=(1+\alpha)\denoiser(\vz_t,t,\vc)-\alpha \denoiser(\vz_t,t,\varnothing).
\end{equation}

Here, $\hat{\rvepsilon}_{\dmparam}$ denotes the classifier free guidance (CFG) version \cite{ho2022classifier} of $\denoiser$ used in text-conditioned DMs to enable higher quality generation via a guidance scale parameter $\alpha$.  Intuitively, CFG aims to trade diversity with quality and is a linear combination of two $\denoiser$ terms, with the condition omitted in the latter (represented here by the null symbol $\varnothing$).



\noindent\textbf{Personalization and Parameter Efficient Finetuning} DMs can be personalized to generate images aligned with a particular concept, subject, or style \cite{ruiz2022dreambooth,shah2023ziplora}, by finetuning their weights on a select number of images. However, updating all the parameters of a pretrained DM is computationally expensive. Instead, parameter-efficient finetuning methods like LoRA \cite{hu2021lora}, inject trainable low-rank decomposition matrices and aim to learn only the variations from the pretrained weights. 

\noindent\textbf{Gaussian Splatting} parameterizes the scene with a set of 3D Gaussian particles $\gG$, where each particle is represented by its position $\bm{\mu} \in \mathbb{R}^3$, scale $\vs \in \mathbb{R}^3$, rotation $\vr \in \mathbb{R}^4$, opacity $\sigma \in \mathbb{R}$, and spherical harmonics coefficients $\bm{\beta} \in \mathbb{R}^{48}$. Gaussian particles can be efficiently rendered using the differentiable splatting formulation proposed in ~\cite{kerbl20233Dgaussians}, and hence, optimized from a set of posed images using a reconstruction loss.
Consequently,  the set of differentiable parameters is defined as $\gparam:=\{\bm{\mu}_i,\vs_i,\vr_i,\sigma_i,\bm{\beta}_i \}_{i=1}^{|\gG|}$.
\vspace{-2mm}
\subsection{Problem Setup}
Given a 3D reconstruction of a scene and the corresponding posed images that generated it, we investigate 3D editing techniques that enable a user to perform certain manipulations. In the following, we detail our contributions on the task of 3D inpainting, before showing results on other applications that can be tackled with the same formulation~\Secref{sec:applications}.
In all the experiments, we adopt Gaussian splatting~\cite{kerbl20233Dgaussians} as our underlying 3D representation. However, our approach can also be applied to other 3D representations such as NeRFs~\cite{Mildenhall2020ECCV}.
\vspace{-2mm}
\subsection{Multi-Scale Personalization}
\label{sec:multiscale_personalization}
Our approach involves distilling the learned priors from 2D generative models into the 3D domain to provide effective generative guidance. Distilling priors from 2D image DMs into 3D using a score distillation objective is commonly recognized as a mode-seeking optimization problem. As a result, the outputs produced by this procedure often exhibit a lack of detail and diversity. In addition, DMs are typically guided using text prompts, which are not suited to describe large-scale scenes and therefore only provide weak regularization. To mitigate these problems, we propose to personalize the DM based on a single (or multiple) reference image of the scene before distilling its priors to 3D. This personalization adapts the DM towards the reference image and alleviates the need for text guidance during the distillation process.

Let $\imgReference \in \mathbb{R}^{h \times w \times 3}$ denote a reference image, which can be generated by an inpainting model or an actual image of the scene. We use LoRA finetuning to personalize a pretrained inpainting LDM to $\imgReference$ by minimizing the objective in~\Eqref{eq:vanilla_difussion}. Specifically, at each iteration, we first augment $\imgReference$ by cropping out a random thin border around it to obtain $\imgReferenceCropped  \in \mathbb{R}^{h' \times w' \times 3}$, where $h' \leq h$ and $w' \leq w$. We then sample a random rectangular mask $\vm \in \{0,1\}^{h' \times w'}$ 
that masks out part of the image and task the LDM to inpaint the missing content. The conditioning $\vc$ in~\Eqref{eq:vanilla_difussion} is obtained by concatenating the latent representation of the masked reference view $\vz=\encoder(\vm \circ \imgReferenceCropped)$ and the mask $\vm$, where $\circ$ denotes the element-wise product. For the text embedding $\vy= \encodertext(C_T)$ we use a fixed prompt $C_T=\globalText$. We apply LoRA to the attention layers of both the text encoder $\encoder_\text{text}$ and the U-Net denoiser $\denoiser$.

In practice, the native image resolution of our LDM is $512 \times 512$ pixels with a $64 \times 64$ dimensional latent space. Such low resolution is often too coarse to capture the high-frequency details of the scene. Especially, as the gradients of the SDS loss are computed in the low-resolution latent space and then backpropagated to the input (\cf \Eqref{eq:sds_gradient}). 
To address this challenge, we propose a multi-scale personalization strategy wherein the diffusion inpainter is additionally finetuned on local $512 \times 512$ crops sampled around the masked region of $\imgReference$. To enable the DM to discern variations between local and global crops we use a distinct text prompt, $C'_T=$\localText~ for these local crops.


\subsection{Multi-Scale SDS Objective}
\label{sec:multiscale_sds}
Building on the multi-scale personalization, we formulate a multi-scale SDS objective:
\begin{equation}
    \mathcal{L}_\text{SDS} := \mathcal{L}_\text{SDS}^\text{global} + \mathcal{L}_\text{SDS}^\text{local} .
    \label{eq:ms_sds}
\end{equation}

Specifically, consider an image $\imgRendered_i$ rendered from a random camera pose from the training set along with its mask, $\vm_i$. Then, $\nabla \mathcal{L}_\text{SDS}^\text{global}$ is computed using the whole $\imgRendered_i$ and the text guidance \globalText, while $\mathcal{L}_\text{SDS}^\text{local}$ considers only a random patch $\imgPatchRendered_i$ around the masked region of $\imgRendered_i$ and is guided using the text prompt $\localText$. Such multi-scale formulation allows us to balance global context with local details even at the low resolution of the DM's latent space.

\subsection{Splitting the Gaussians into Masked/Unmasked Sets}
\label{sec:splitting_the_gaussians}
We assume that the region of the scene that should be removed and inpainted is either masked directly in 3D or provided in the form of 2D masks for each training view. The latter can be generated using an off-the-shelf 2D segmentation method or provided by the user. As such 2D masks can be inconsistent in 3D, we propose a simple heuristic that consolidates them.

\noindent\textbf{3D Consistent Masks}
To consolidate inconsistent 2D masks, we first determine if each Gaussian particle should be masked or unmasked, by counting how often it contributes to the volume rendering of masked and unmasked pixels across the training views. If the ratio masked/unmasked is above a threshold $\tau_\text{mask}$ the 3D particle is labeled as masked. After particle labelling, we rasterize them back to the training views and threshold the rendered semantic values using $\tau'_\text{masked}$, producing a set of 3D consistent 2D binary masks.

\noindent\textbf{Confining the Loss Functions to Pertinent Regions}
During the optimization, we use the per-Gaussian masks to direct the gradients of individual loss terms. Specifically, the reconstruction loss from unmasked pixels only updates the parameters of unmasked Gaussians, whereas the gradients from other loss terms are only propagated to the masked Gaussians. This differentiation into masked and unmasked Gaussians is crucial to prevent guidance losses (e.g., SDS loss) from undesirably influencing the regions that do not require hallucination. But, it also requires us to adapt the densification and pruning heuristics proposed in~\cite{kerbl20233Dgaussians}. Specifically, if a Gaussian is split or cloned we transfer its mask value to its children. Additionally, if a Gaussian transitions between the regions, we prune it to maintain region-specific fidelity. 

\noindent\textbf{Reference-Guided Initialization}
At the start of the inpainting process, we remove  Gaussians in the masked scene region and replace them using the reference view $\imgReference$. Specifically, we first predict the depth $\tilde d_r$ of the reference image $\imgReference$. We then compensate the scale and offset ambiguities to derive the aligned depth $\hat d_r$ (similar to what will follow in our depth regularization) and unproject it to 3D. We empirically observe that reference-guided initialization outperforms random initialization~\tabref{ablation_study}.

\subsection{Losses and Training}
We train our method using the combination of the multiscale SDS objective~\Secref{sec:multiscale_sds} and the reconstruction loss $\mathcal{L}_\text{rec}$ ($L1$ and D-SSIM). Additionally, we employ two regularization terms that improve the geometry and appearance of the synthesized region. The overall objective is:
\begin{equation}
    \label{eq:overall_objective}
    \mathcal{L} := 
    \mathcal{L}_\text{rec} + 
    \lambda_\text{SDS} \mathcal{L}_\text{SDS} + 
    \lambda_\text{depth} \mathcal{L}_\text{depth} + 
    \lambda_\text{adv} \mathcal{L}_\text{adv},
\end{equation}
where $\lambda_\ast > 0$ balance the relative contribution of each loss term and $\mathcal{L}_\text{adv}$ and $\mathcal{L}_\text{depth}$ denote the adversarial and depth loss, respectively.

\noindent\textbf{Depth Regularization}
We regularize the depth using the outputs of a monocular depth estimation model. In particular, after every $N_\text{depth}$ iterations, we render an image $\hat{\mathbf{I}}_i$ from a randomly selected camera pose in the training set. We then mask $\hat{\mathbf{I}}_i$ with its corresponding mask $\vm_i$ and inpaint it using our adapted inpainting model starting from a random timestep $t_\text{depth} \in T$. This process yields an inpainted view $\hat{\mathbf{I}}_\text{inpaint}$ that we feed into a monocular depth estimation model to obtain the depth map $\tilde d$. 

Note that $\tilde d$ is only relative and ambiguous in terms of scale and offset. To fix the ambiguity, we optimize a scale $s$ and offset $o$ to minimize the $L2$ error between the rendered depth, $\hat d$, and the aligned depth $\bar d = s \tilde d + o$ for the set of unmasked pixels $P_\text{unmasked}$. Finally, we employ the fast bilateral solver~\cite{barron2016fast} to reduce potential remaining misalignments and calculate $\mathcal{L}_\text{depth}$ on the masked pixels, $P_\text{masked}$, as:
\begin{equation}
    \mathcal{L}_\text{depth} := \frac{1}{\vert P_\text{masked} \vert} \sum_{p\in P_\text{masked}} \Vert \hat d(p) - \bar d(p) \Vert_2^2.
\end{equation}

\noindent\textbf{Adverserial Objective}
To mitigate any color mismatches and artifacts on the boundary of the hallucinated region, we employ a discriminator $\mathcal{D}_{\bm{\xi}}$ parameterized by $\bm{\xi}$. This discriminator is trained to differentiate between real patches that are sampled around the masked region of the reference view $\imgReference$ and the fake patches rendered from 
$\gnet(\vpi)$
around the mask of the training views. We propagate the gradients of the adversarial loss to all spherical harmonics coefficients $\bm{\beta}_i \in \gparam$, while keeping other parameters fixed.
Following GaNeRF~\cite{roessle2023ganerf}, we employ the following regularized version of the GAN loss $\mathcal{L}_\text{adv}$ with an $R_1$ gradient penalty~\cite{mescheder2018training} on 
$\mathcal{D}_{\bm{\xi}}$, 
controlled by a balancing scalar $\lambda_\text{gp}$:
\begin{equation}
    \label{eq:gan_loss}
    \resizebox{0.9\hsize}{!}{
    $\text{min}_{\bm{\beta}}\text{max}_\mathbb{\xi} \mathbb{E} \Big[  
        f(\mathcal{D}_\mathbb{\xi}(\mathbf{\hat{I}}^P_\text{fake})) + f(-\mathcal{D}_\mathbb{\xi}(\mathbf{I}^P_\text{real})) - \lambda_\text{gp} \Vert \nabla \mathcal{D}_\mathbb{\xi}(\mathbf{\hat{I}}^P_\text{fake}) \Vert_2^2
    \Big],$
    }
\end{equation}
where $f(x):= -\log (1 + \exp (-x))$, and $\mathbf{\hat{I}}^P_\text{fake}$ and $\mathbf{I}^P_\text{real}$ correspond to the sampled fake and real patches, respectively.

\subsection{Implementation Details}
\label{sec:implementation_details}
For unprojection and reprojection of the masks, we set $\tau_\text{masked}$ and $\tau'_\text{masked}$ to $1$ and $0.3$, respectively. 
In each iteration, we compute $\mathcal{L}_\text{SDS}^\text{global}$ using an image rendered from a random camera pose from the training set, while averaging the $\mathcal{L}_\text{SDS}^\text{local}$ across two $512\times 512$ patches sampled from the bounding box of the masked region. Following Gaussian splatting~\cite{kerbl20233Dgaussians}, we resize each training view to have the larger side of the images be $1600$ pixels. The depth loss is calculated every 8th iteration. For the adversarial loss, we sample $64$ real and $64$ fake $64\times 64$ patches every iteration. The architecture of the discriminator follows  StyleGAN2~\cite{karras2020stylegan2}, and the hyperparameters of the adversarial loss are consistent with GANeRF~\cite{roessle2023ganerf}. The loss weights of individual loss terms are set to $\lambda_\text{SDS} := 0.001$, $\lambda_\text{depth}:=0.0625$, and $\lambda_\text{adv} := 0.03$, respectively.  We adopt the Gaussian optimization parameters from \cite{kerbl20233Dgaussians}, but change the \textit{densification and cloning} heuristic to prevent overdensification of Gaussians due to larger gradients of the SDS loss. To generate the refence views we use SDXL~\cite{podell2023sdxl} inpainting model due to its higher resolution and quality. However, to speed up the distillation, we personalize Stable-Diffusion-2-Inpainting~\cite{Rombach_2022_CVPR} as the adapted LDM. The same LoRA optimization parameters as RealFill~\cite{tang2023realfill} are employed; we train the DM for 2000 iterations, with LoRA rank set to $8$, $\alpha$ set to $16$, resolution set to $512$, and learning rates $2e^{-4}$ and $4e^{-5}$ for the U-Net and the text encoder, respectively. We set the LoRA dropout to $0.1$. Note that the adapted LDM expects $512\times 512$ images; thus, for calculating the global SDS loss, we first bilinearly downsample the rendered views to $512 \times 512$.

\section{Experiments}
\label{sec:experiments}

\noindent\textbf{Dataset and Metrics} 
Following related work, we perform most experiments on the SPIn-NeRF~\cite{mirzaei2023spin} dataset. It consists of 10 scenes originally designed for object removal evaluation. Each scene in the dataset includes 60 images with an unwanted object (training views) and 40 images without it (test views). Human-annotated masks of the object region are available for both training and test views. For more information about the dataset please refer to~\cite{mirzaei2023spin}. To perform a quantitative comparison of our method to the selected baselines, we use 40 ground-truth images of each scene without the object. Specifically, we render corresponding views from each model and compute the average learned perceptual image patch similarity (LPIPS)~\cite{zhang2018perceptual}\footnote{Prior work also reports per-scene FID score, but due to the small size of the dataset (40 images) FID values are unrepresentative. We thus omit them and only report LPIPS.}. In line with the related work~\cite{mirzaei2023spin} we calculate the metrics around the masked region by considering the bounding box of the mask and dilating the box in every direction by 10\%. To address the limitations of SPIn-NeRF dataset in demonstrating method behaviors on scenes with more significant camera motion for inpainting, we introduce our dataset comprising nine scenes. This dataset encompasses wide-baseline forward-facing scenes as well as 360-degree scenes. For each scene, our dataset includes images both with and without unwanted objects, along with human-annotated masks for every scene.

We further performed a user study using Amazon Mechanical Turk. Specifically, raters were shown videos depicting the results of two methods rendered from the same novel trajectories (our method and a baseline in random order), along with the input video and images highlighting the unwanted object. Raters were then asked to vote for the method that produced the best video in terms of \emph{overall quality} and best \emph{object removal}. To improve the quality of responses, we introduced an attention check task. For each scene, we added a question where the videos displayed for both methods were the same (outputs of our model) and added an option suggesting that both of the videos are identical. We set a threshold of 50\% accuracy on the attention check questions to strike a balance between the quantity and quality of the responses. In total, this resulted in 32 users comparing our method to three baselines across 10 scenes and answering 2 questions per example. 

\noindent\textbf{Baselines}
We compare \MethodName to two naive baselines, \textit{Masked NeRF} and \textit{NeRF + LaMa}. In the former, the reconstruction loss is only calculated on the unmasked pixels, while the latter uses LaMa~\cite{suvorov2022resolution} to independently inpaint rendered views. We further compare to a wide range of existing 3D inpainting methods: ObjectNeRF~\cite{yang2021objectnerf}, NeRF-In~\cite{liu2022nerfin}, SPIn-NeRF~\cite{mirzaei2023spin}, and concurrent work Inpaint3D~\cite{prabhu2023inpaint3d}. For all the baselines, we used the source code or the rendered images provided by the authors of the respective works.

\begin{figure*}
    \centering
   \includegraphics[width=0.99\textwidth]{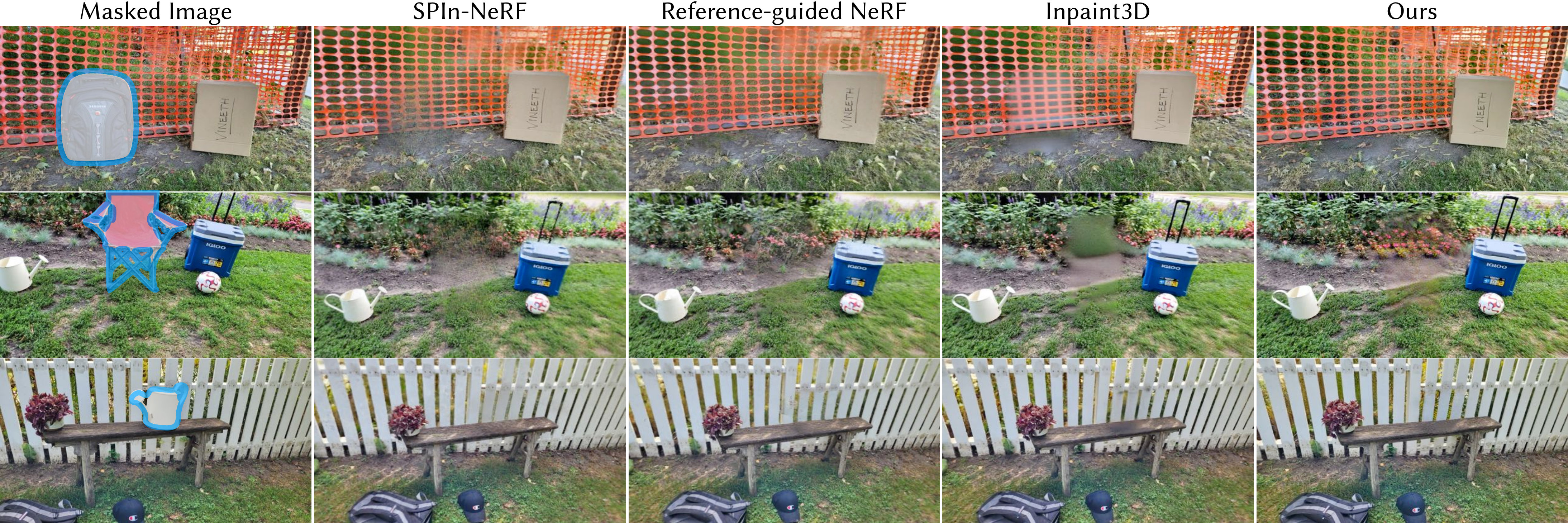}
   \vspace{-3mm}
    \caption{Qualitative object removal results on the SPIn-NeRF dataset. \MethodName consistently outperforms the baselines, yielding sharper reconstruction and more plausible inpainting.}
    \label{fig:baseline_comparison}
    \vspace{-3mm}
\end{figure*}

\noindent\textbf{Object Removal} 
We first provide evaluations based on the standard 3D inpainting benchmark, the SPIn-NeRF dataset~\cite{mirzaei2023spin}. To ensure a fair comparison, we use the human-annotated masks from the SPIn-NeRF dataset for all methods. As depicted in~\tabref{qualitative_evaluation} our method surpasses all baselines in terms of LPIPS. Furthermore, we also compare favourably in a qualitative comparison as seen in ~\Figref{fig:baseline_comparison}. Note that the perceptual loss used in SPIn-NeRF doesn't fully resolve the multi-view inconsistencies introduced by inpainting each training view separately, therefore resulting in visual artifacts. Reference-guided NeRF performs well in the removal region, however the quality of the overall reconstruction quality suffers. Finally, Inpaint3D also distills a pretrained 2D inpainting DM using an SDS objective, however the lack of personalization results in a fuzzy inpainting. We summarize the results of the user study in ~\tabref{user_study}, \MethodName outperforms all baselines in terms of \emph{overall quality} and \emph{object removal}.

\begin{table}[]
\caption{Quantitative evaluation of object removal on SPIn-NeRF dataset.}
\vspace{-2mm}
\begin{tabular}{ll}
\hline
Method                                              & LPIPS$\downarrow$  \\ \hline
NeRF + LaMa (2D)                                    &  0.5369               \\
ObjectNeRF~\cite{yang2021objectnerf}                &  0.6829               \\
Masked NeRF~\cite{Mildenhall2020ECCV}                                 &  0.6030               \\
NeRF-In~\cite{liu2022nerfin}                          &  0.4884               \\
SPIn-NeRF-SD~\cite{mirzaei2023spin}                 &  0.5701               \\
SPIn-NeRF-LaMa~\cite{mirzaei2023spin}               &  0.4654               \\
Inpaint3D~\cite{prabhu2023inpaint3d}                &  0.5150               \\
Reference-guided NeRF~\cite{mirzaei2023reference} (SDV2)   &  0.4532             \\
Reference-guided NeRF~\cite{mirzaei2023reference} (SDXL)   &  0.4453             \\
\hline
Ours                                                &  \textbf{0.4283}      \\ \hline
\end{tabular}
\label{tab:qualitative_evaluation}
\vspace{-3mm}
\end{table}

\begin{table}[]
\caption{User study of object removal on SPIn-NeRF dataset. For each method we report the percentage of raters that pereferred it over ours.}
\vspace{-2mm}
\begin{tabular}{lcc}
\hline
Method                                                     & Quality (\%) & Removal (\%) \\ \hline
SPIn-NeRF-LaMa                 &  18.38 & 29.32   \\
Inpaint3D                      &  11.87 & 11.52   \\
Reference-guided NeRF (SDXL)   &  23.64 & 43.65   \\ \hline
\end{tabular}
\label{tab:user_study}
\vspace{-2mm}
\end{table}

\begin{figure*}
    \centering
   \includegraphics[width=0.99\textwidth]{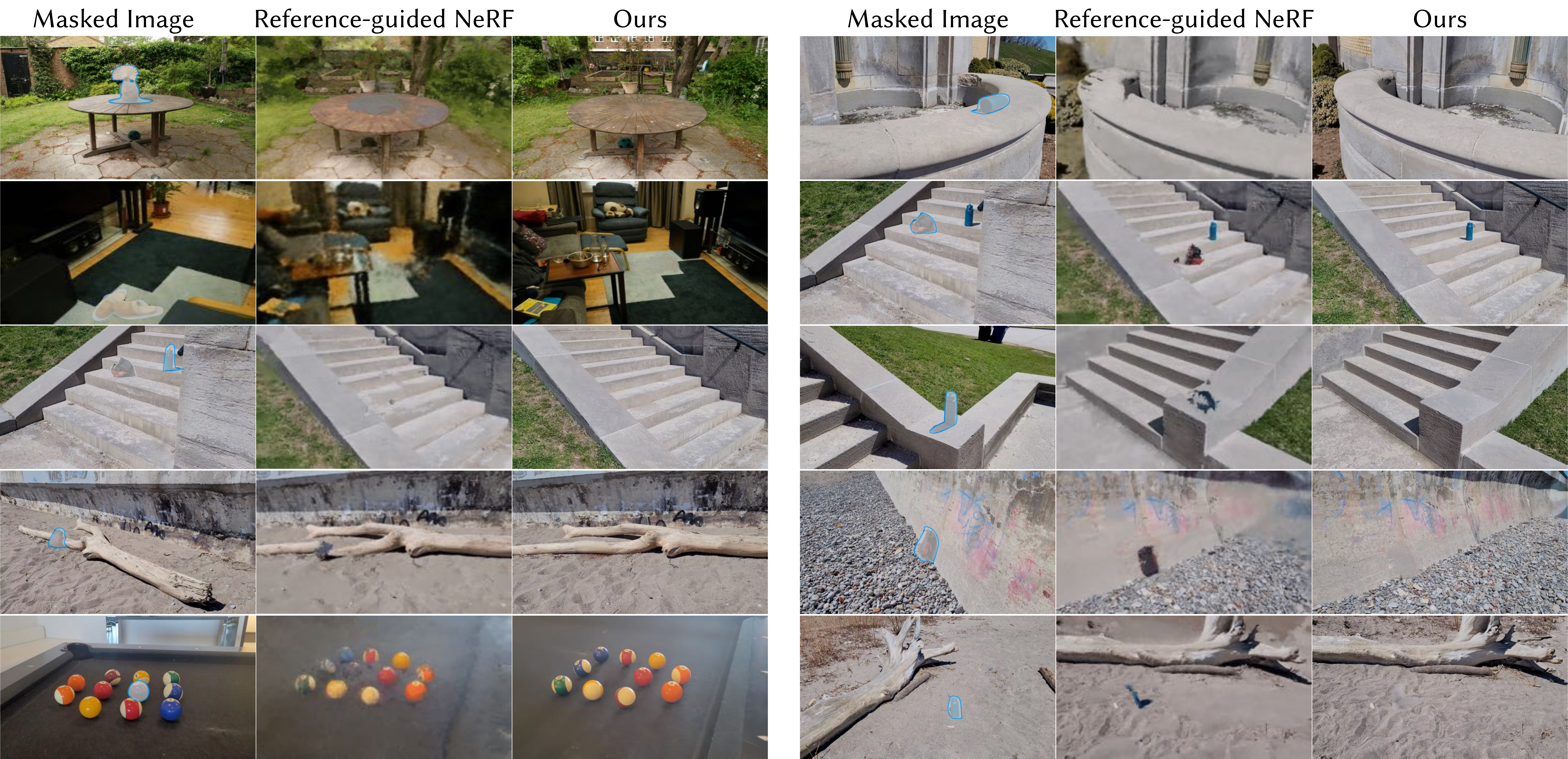}
   \vspace{-3mm}
    \caption{Qualitative object removal results on scenes with larger camera movements (MipNeRF360 dataset~\cite{barron2022mipnerf360} and scenes from our proposed dataset). \MethodName consistently outperforms the Reference-guided NeRF.}
    \label{fig:wide_baseline_rgnerf_comparison}
    \vspace{-2mm}
\end{figure*}

The SPIn-NeRF dataset~\cite{mirzaei2023spin} exclusively comprises scenes with minimal camera movements. To remedy this constraint, we introduce our dataset featuring nine distinct scenes with broader camera baselines, including wide-baseline forward-facing scenes and 360-degree scenes. In \Figref{fig:wide_baseline_rgnerf_comparison}, we present qualitative comparisons between the outcomes of our approach and Reference-guided NeRF applied to scenes from the MipNeRF360 dataset~\cite{barron2022mipnerf360} (depicted in the first and second scenes of the left column), as well as our own scenes. The results notably illustrate that Reference-guided NeRF, reliant solely on a single inpainted view for projection into 3D space, encounters challenges in extrapolating information from the reference to further views. Conversely, our method consistently yields sharper and more plausible inpainting results without visual artifacts. Note that in the third example, we remove two distinct objects by concurrently incorporating both into the masks.

\noindent\textbf{Ablation on Design Choices} We perform ablation studies on our key design choices to highlight their significance. ~\tabref{ablation_study} and ~\Figref{fig:ablation_study} depict quantitative and qualitative results. The largest drop in performance is observed when removing \emph{personalization} from our method. This confirms our intuition that reference adaptation is crucial for unlocking the potential of SDS-based optimization at the scene level. Other ablations show that removing any components leads to adverse effects on the final outcomes. While we found the depth loss not to be crucial for the object removal task, it still contributes to a subtle improvement. Moreover, it is particularly beneficial for other applications such as object insertion.

\begin{table}[]
\caption{Ablation study of object removal on SPIn-NeRF dataset.}
\vspace{-2mm}
\begin{tabular}{ll}
\hline
Method                                              & LPIPS$\downarrow$     \\ \hline
Ours w/o personalization                            &  0.5719               \\
Ours w/o splitting the Gaussians                    &  0.5128               \\
Ours w/o local SDS                                  &  0.5093               \\
Ours w/o reference-guided initialization            &  0.4680               \\
Ours w/o adversarial loss                           &  0.4326               \\
Ours w/o depth loss                                 &  0.4299               \\
\hline
Ours                                                &  \textbf{0.4283}      \\ 
\hline
\end{tabular}
\label{tab:ablation_study}
\vspace{-3mm}
\end{table}

\section{Applications}
\label{sec:applications}

In this section, we demonstrate the general applicability of our method to various downstream 3D applications. 

\begin{figure}
    \centering
   \includegraphics[width=0.47\textwidth]{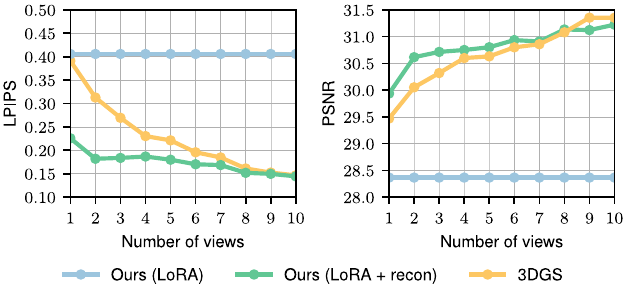}
   \vspace{-4mm}
    \caption{Results of the sparse view reconstruction on SPIn-NeRF dataset. Using the sparse GT views only for personalization \emph{Ours (LoRA)} already yields competitive results. When combined with the reconstruction loss \emph{Ours (LoRA + recon)} consistently outperforms 3DGS~\cite{kerbl20233Dgaussians}, showcasing the potential of generative priors to guide 3D reconstruction.}
    \label{fig:gt_experiment}
\end{figure}

\noindent\textbf{Sparse View Reconstruction}
To investigate the benefits of our method for guiding the sparse view reconstruction, we consider a scene with an unwanted occluder where only a small set of clean images from the scene (GT views) is available. Specifically, we consider two settings: a) we only use the GT views for LoRA finetuning and do not use them during the reconstruction---\textit{Ours (LoRA)}, or b) we additionally use the GT views for supervision using reconstruction loss---\textit{Ours (LoRA + recon)}. ~\Figref{fig:gt_experiment} shows that generative priors can guide the reconstruction in a sparse view setting\footnote{A similar observation was concurrently shown by \cite{wu2023reconfusion}.}, especially when only a small number of views are available. Indeed, \textit{Ours (LoRA + recon)} consistently outperforms 3DGS in terms of LPIPS and PSNR. \Figref{fig:gt_experiment_qualitative} depicts a qualitative comparison with 3DGS.




\begin{figure}
    \centering
   \includegraphics[width=0.46\textwidth]{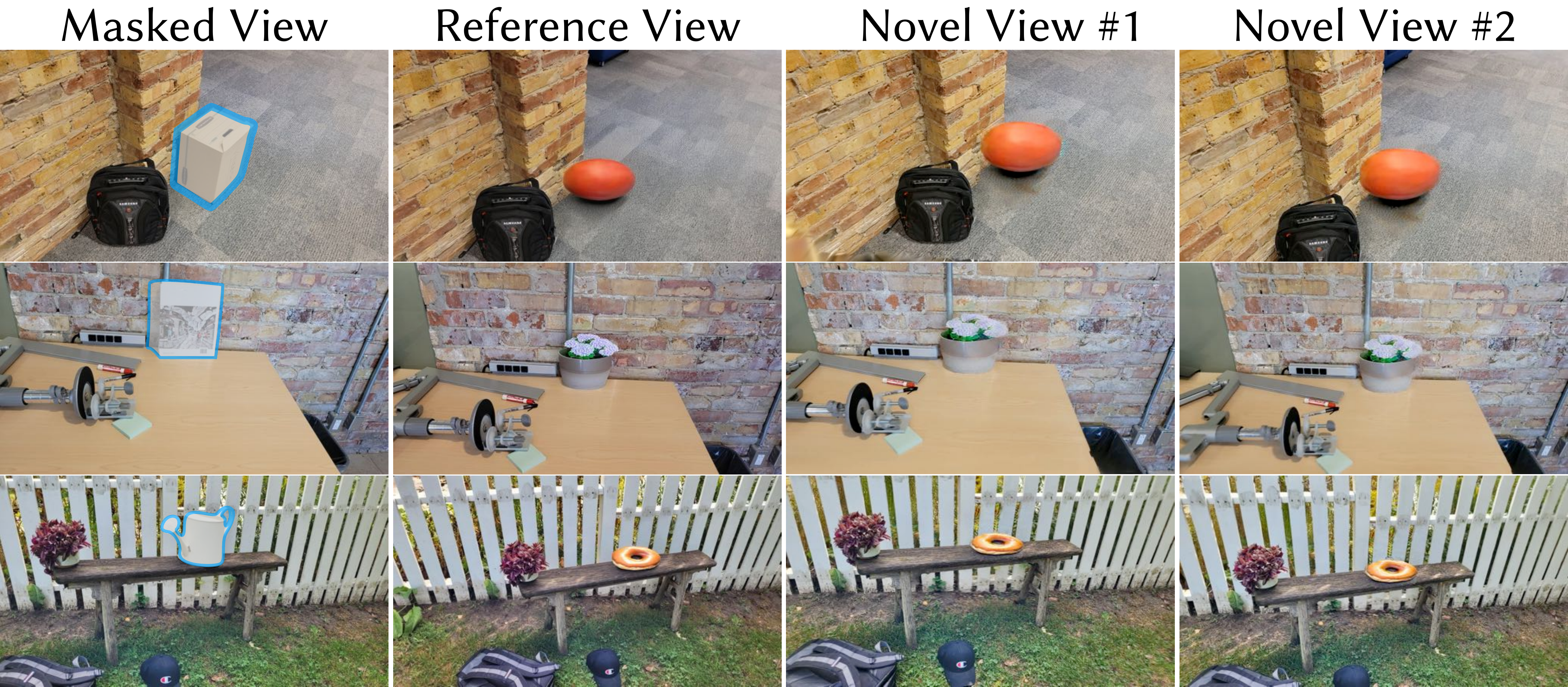}
   \vspace{-3mm}
    \caption{Sample object insertion results.}
    \label{fig:different_references}
    \vspace{-3mm}
\end{figure}

\noindent\textbf{Object Insertion}
\Figref{fig:different_references} illustrates the capacity of \MethodName to insert objects into a scene. We demonstrate that by using a reference view with an added object in the masked region, obtained using a text-to-image inpainting diffusion model. Our method succeeds in distilling the specified object into the scene with high visual fidelity. 

\begin{figure}
    \centering
   \includegraphics[width=0.46\textwidth]{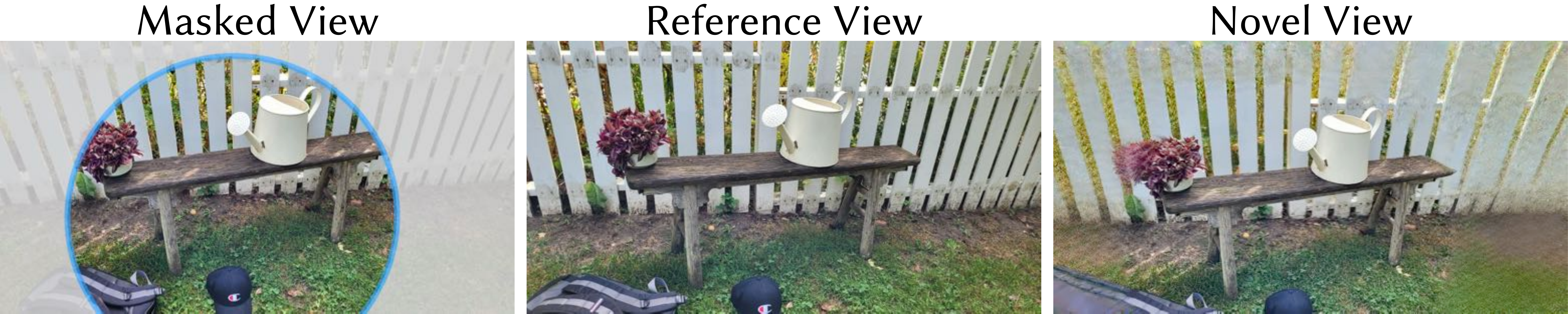}
   \vspace{-3mm}
    \caption{Our approach is capable of outpainting scenes by inverted masks.}
    \vspace{-4mm}
    \label{fig:outpainting}
\end{figure}

\noindent\textbf{Scene Outpainting}
In~\Figref{fig:outpainting}, we follow the procedure proposed in~\cite{prabhu2023inpaint3d} and generate inverse masks by placing a sphere at a fixed distance along the optical axis and checking for ray-sphere intersection. Example mask is shown in ~\Figref{fig:outpainting} (left). Given this mask, we task our method to outpaint the scene using the same formulation and hyperparameters used for object removal. Our method completes the scene in a plausible manner, but the outpainted regions lack visual fidelity and high-frequency details. This hints that special treatment is required to obtain the same quality of results when performing outpainting. 

\section{Conclusions, Limitations and Future Work}
\label{sec:conclusion}

We introduced \MethodNameFull,  a 3D inpainting framework that achieves state-of-the-art results, while offering improved controllability over the inpainted content. We achieve this by personalizing an inpaitning LDM to the given reference image and distilling the adapted diffusion prior to the 3D scene We demonstrate the generality of our method across a diverse set of downstream applications including object insertion, 3D outpainting, and sparse view reconstruction.

Despite achieving notable results, there are many avenues to improve 
\MethodNameFull. The primary challenge lies in removing large objects that substantially cover parts of the reference image. Finetuning the diffusion model on the target dataset could help bridge this gap. Additionally, although LoRA improves the efficiency of the adaptation process to the target scene,  this process still requires a considerable amount of time. Further advancements in parameter-efficient finetuning and personalization of diffusion models are expected to further speed up and enhance our approach.

Finally, recent advances in multi-view aware and video diffusion models could significantly enhance the multi-view quality of our results. Especially, as the remaining artifacts are mostly visible when the camera moves.


{\small
\bibliographystyle{ACM-Reference-Format}
\bibliography{egbib}
}

\appendix
\begin{figure*}
    \centering
   \includegraphics[width=0.97\textwidth]{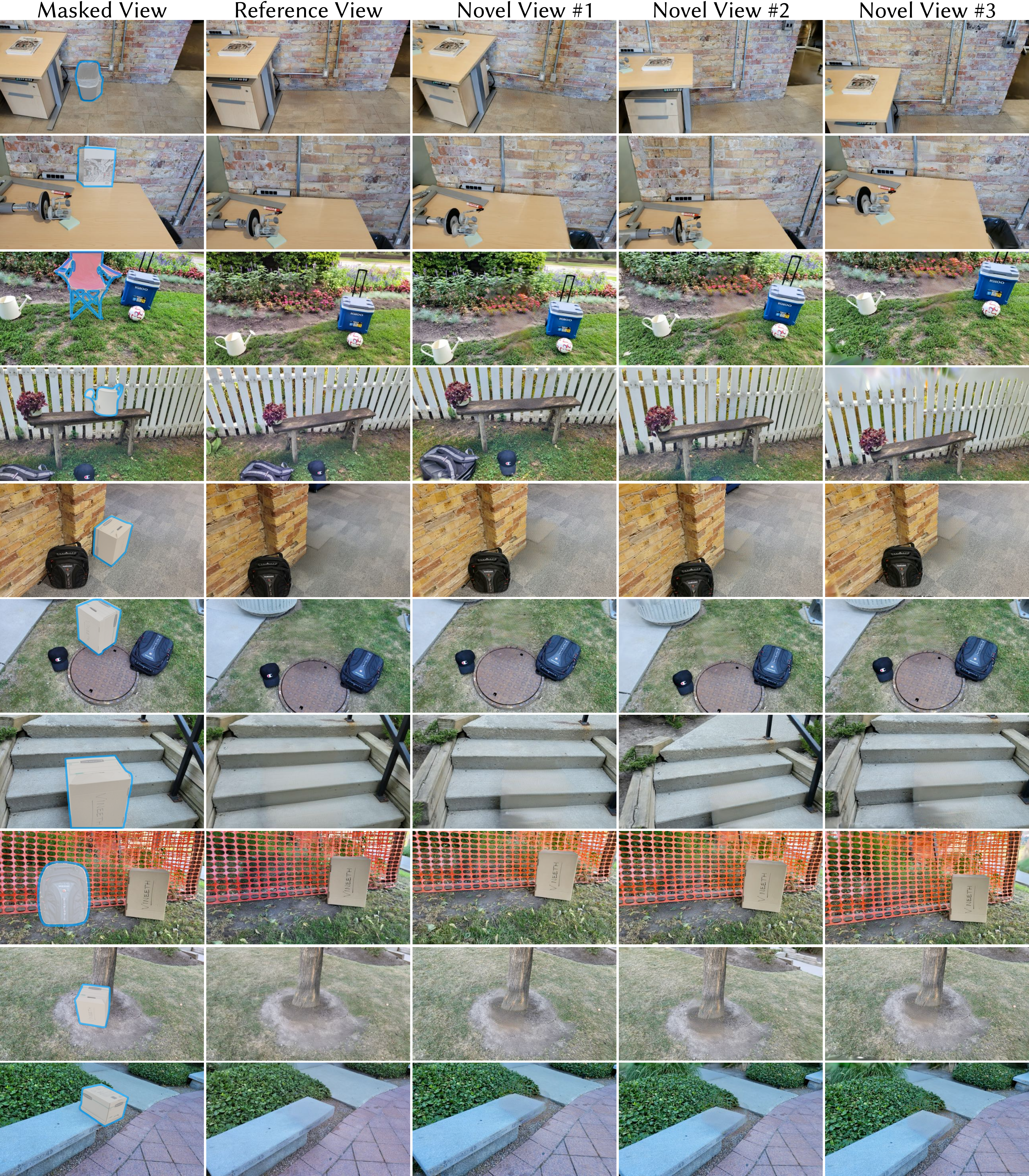}
    \caption{Qualitative object removal results on the SPIn-NeRF dataset. \MethodName synthesizes plausible content that is highly multi-view consistent.}
    \label{fig:qualitative_results_spinnerf}
\end{figure*}

\begin{figure*}
    \centering
   \includegraphics[width=\textwidth]{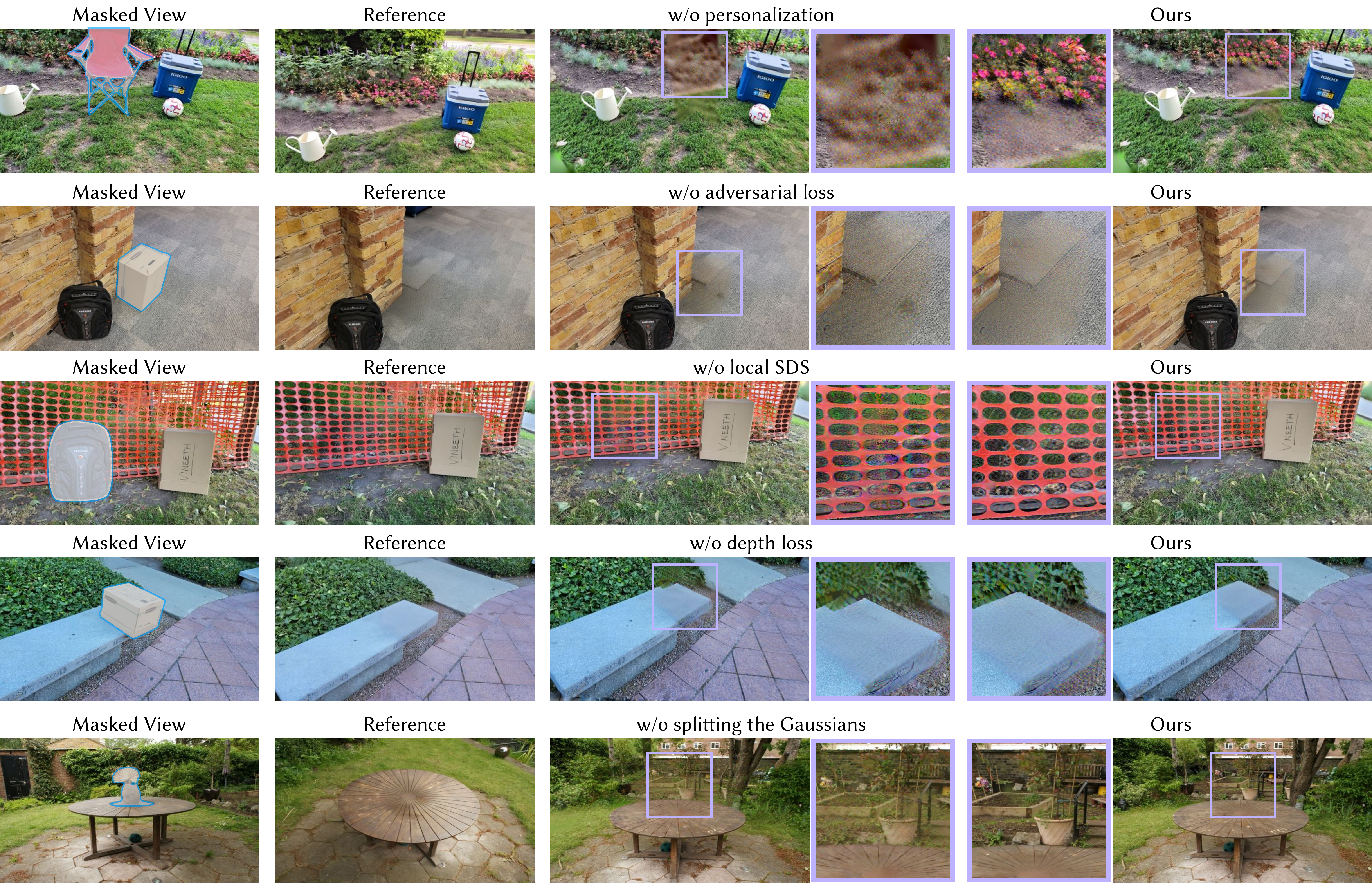}
    \caption{Qualitative results of the ablation study on SPIn-NeRF dataset. Note how different components of our method help improve different types of artifacts.}
    \label{fig:ablation_study}
\end{figure*}

\begin{figure*}
    \centering
   \includegraphics[width=1\textwidth]{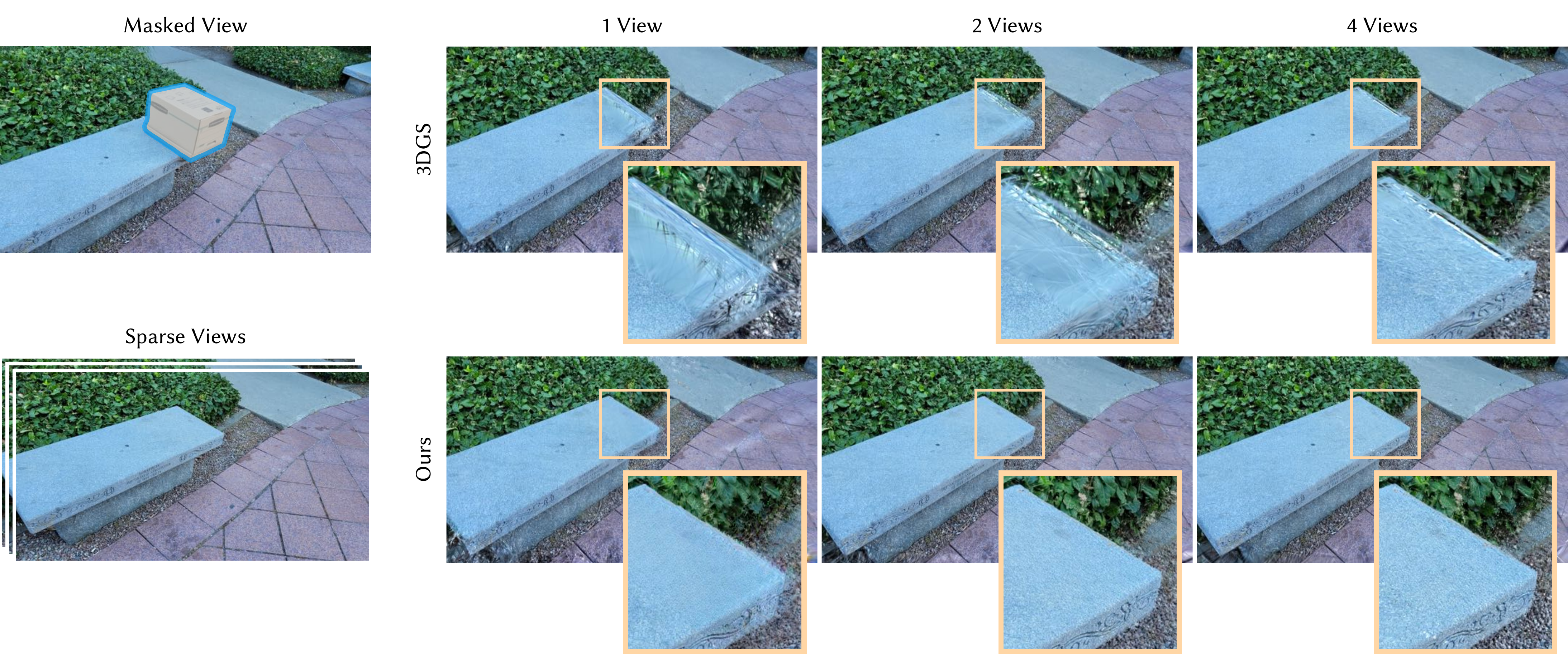}
    \caption{Qualitative evaluation of sparse view reconstruction on SPIn-NeRF dataset. Both \MethodName and 3DGS use the reconstruction loss on sparse input images in the masked region. Additionally, \MethodName uses generative priors of the reference adapted LDM through SDS losses as well as the depth and adversarial regularization terms. Note how generative priors can successfully guide the reconstruction even in the extreme case of a single GT view.}
    \label{fig:gt_experiment_qualitative}
\end{figure*}

\end{document}